\documentclass{article}

\usepackage{times}
\usepackage{graphicx} 
\usepackage{subcaption}
\usepackage{amsfonts}
\usepackage{amsmath}
\usepackage{bm}
\usepackage[disable]{todonotes} 

\usepackage{tikz}
\usetikzlibrary{calc}
\def\tikzmark#1{\tikz[remember picture,overlay]\coordinate(#1);}

\usepackage{natbib}

\usepackage{algorithm}
\usepackage{algorithmic}

\usepackage{hyperref}

\usepackage[accepted]{icml2017} 

\icmltitlerunning{Parallel and Distributed Thompson Sampling for Large-scale Accelerated Exploration of Chemical Space}

\begin{document} 

\twocolumn[
\icmltitle{Parallel and Distributed Thompson Sampling for\\ Large-scale Accelerated Exploration of Chemical Space}



\icmlsetsymbol{equal}{*}

\begin{icmlauthorlist}
\icmlauthor{Jos{\'{e}} Miguel Hern{\'{a}}ndez{-}Lobato}{equal,cam}
\icmlauthor{James Requeima}{equal,cam,inv}
\icmlauthor{Edward O. Pyzer-Knapp}{har,ibm}
\icmlauthor{Al{\'{a}}n Aspuru-Guzik}{har}
\end{icmlauthorlist}

\icmlaffiliation{cam}{University of Cambridge, Cambridge, UK}
\icmlaffiliation{har}{Harvard University, Cambridge, USA}
\icmlaffiliation{ibm}{IBM Research, UK}
\icmlaffiliation{inv}{Invenia Labs, Cambridge, UK}

\icmlcorrespondingauthor{Jos{\'{e}} Miguel Hern{\'{a}}ndez{-}Lobato}{jmh233@cam.ac.uk}
\icmlcorrespondingauthor{Edward O. Pyzer-Knapp}{epyzerk3@uk.ibm.com}

\icmlkeywords{machine learning, Thompson sampling, Bayesian optimization, ICML}

\vskip 0.3in
]



\printAffiliationsAndNotice{\icmlEqualContribution} 

\begin{abstract}
\todo{I'd like a motivating sentence here.}
Chemical space is so large that brute force searches for new interesting
molecules are infeasible. \emph{High-throughput virtual screening} via computer cluster simulations can speed up
the discovery process by collecting very large amounts of data in parallel,
e.g., up to hundreds or thousands of parallel measurements. \todo{Explain the ``virtual'' aspect.} Bayesian
optimization (BO) can produce additional acceleration by sequentially
identifying the most useful simulations or experiments to be performed next. \todo{Talk about optimal experimental design?}
However, current BO methods cannot scale to the large numbers of parallel
measurements and the massive libraries of molecules currently used in
high-throughput screening. Here, we propose a scalable solution based on a
parallel and distributed implementation of Thompson sampling (PDTS). We show
that, in small scale problems, PDTS performs similarly as parallel expected improvement (EI), a batch version of the most
widely used BO heuristic. Additionally, in settings where parallel EI does not scale,
PDTS outperforms other scalable baselines such as a greedy search,
$\epsilon$-greedy approaches and a random search method. These results show that PDTS
is a successful solution for large-scale parallel BO.
\end{abstract} 

\vspace{-0.20cm}
\section{Introduction}

\vspace{-0.10cm}

\todo{Need motivating sentence.}
Chemical space is huge: it is estimated to contain over $10^{60}$ molecules.
Among these, fewer than 100 million compounds can be found in public
repositories or databases \cite{Reymond_2012}. This discrepancy between
\textit{known} compounds and \textit{possible} compounds
indicates the potential for discoverying many new compounds with highly desirable functionality
(e.g., new energy materials, pharmaceuticals, dyes, etc.). While the
vast size of chemical space makes this an enormous opportunity, it also
presents a significant difficulty in the identification of new relevant
compounds among the many unimportant ones. This challenge is so great that any
discovery process relying purely on the combination of scientific intuition
with trial and error experimentation is slow, tedious and in many cases
infeasible.

To accelerate the search, high-throughput approaches can be used in a
combinatorial exploration of small specific areas of chemical space
\cite{Rajan_2008}. These have led to the development of high-throughput virtual
screening \cite{Pyzer_Knapp_2015,G_mez_Bombarelli_2016} in which large
libraries of molecules are created and then analyzed using theoretical and
computational techniques, typically by running a large number of parallel
simulations in a computer cluster. The objective is to reduce an initially very
large library of molecules to a small set of promising leads for which expensive experimental evaluation is justified.  However, even though these techniques only search
a tiny drop in the ocean of chemical space, they can result in massive
libraries
whose magnitude exceeds traditional computational capabilities. As a result,
at present, there is an urgent need to accelerate high-throughput screening
approaches.

Bayesian optimization (BO) \cite{jones1998efficient} can speed up the discovery
process by using machine learning to guide the search and make improved
decisions about what molecules to analyze next given the data collected so far. \todo{This really needs to connect with the larger literature on Bayesian optimal experimental design.}
However, current BO methods cannot scale to the large number of parallel
measurements and the massive libraries of candidate molecules currently used in
high-throughput screening \cite{Pyzer_Knapp_2015}. \todo{Explain why BO doesn't scale.} While there are BO methods that
allow parallel data collection, these methods have typically been limited to tens of data points per batch
\cite{snoek2012practical,shahriari2014entropy,GonDaiHenLaw16}. In contrast,
high-throughput screening may allow the simultaneous collection of thousands
of data points via large-scale parallel computation. This creates a need for new scalable methods for parallel Bayesian optimization.

To address the above difficulty, we present here a scalable solution for
parallel Bayesian optimization based on a distributed implementation of the
Thompson sampling heuristic \cite{Thompson_1933,Chapelle2011}. We show that,
for the case of small batch sizes, the proposed parallel and distributed
Thompson sampling (PDTS) method performs as well as a parallel implementation
of expected improvement (EI) \cite{snoek2012practical,ginsbourger2011dealing},
the most widely used Bayesian optimization heuristic. Parallel EI selects the batch entries sequentially and so EI proposals can't be parallelized, which limits its scalability properties. PDTS generates each batch of evaluation locations by selecting the
different batch entries independently and in parallel. Consequently, PDTS is highly scalable and applicable to large batch sizes.
We also evaluate the performance of PDTS in several real-world high-throughput screening
experiments for material and drug discovery, where parallel EI is infeasible. In these problems, PDTS
outperforms other scalable baselines such as a greedy search strategy,
$\epsilon$-greedy approaches and a random search method. 
These results indicate
that PDTS is a successful solution for large-scale parallel Bayesian
optimization.


\begin{algorithm}
   \caption{Sequential Thompson sampling} \label{alg:seq_thompson_sampling}
\begin{algorithmic}
   \STATE {\bfseries Input:} initial data $\mathcal{D}_{\mathcal{I}(1)}=\{ (\mathbf{x}_i, y_i) \}_{i\in \mathcal{I}(1)}$
   \FOR{$t=1$ {\bfseries to} $T$}
   \STATE Compute current posterior $p(\bm \theta|\mathcal{D}_{\mathcal{I}(t)})$
   \STATE Sample $\bm \theta$ from $p(\bm \theta|\mathcal{D}_{\mathcal{I}(t)})$
   \STATE Select $k\leftarrow \text{argmax}_{j \not \in {\mathcal{I}(t)}} \mathbf{E}[y_j|\mathbf{x}_j,\bm \theta]$
   \STATE Collect $y_k$ by evaluating $f$ at $\mathbf{x}_k$
   \STATE $\mathcal{D}_{\mathcal{I}(t+1)}\leftarrow \mathcal{D}_{\mathcal{I}(t)}\cup \{(\mathbf{x}_k,y_k)\}$
   \ENDFOR
\end{algorithmic}
\end{algorithm}

\begin{figure}
\centering
\includegraphics[width=0.9\columnwidth]{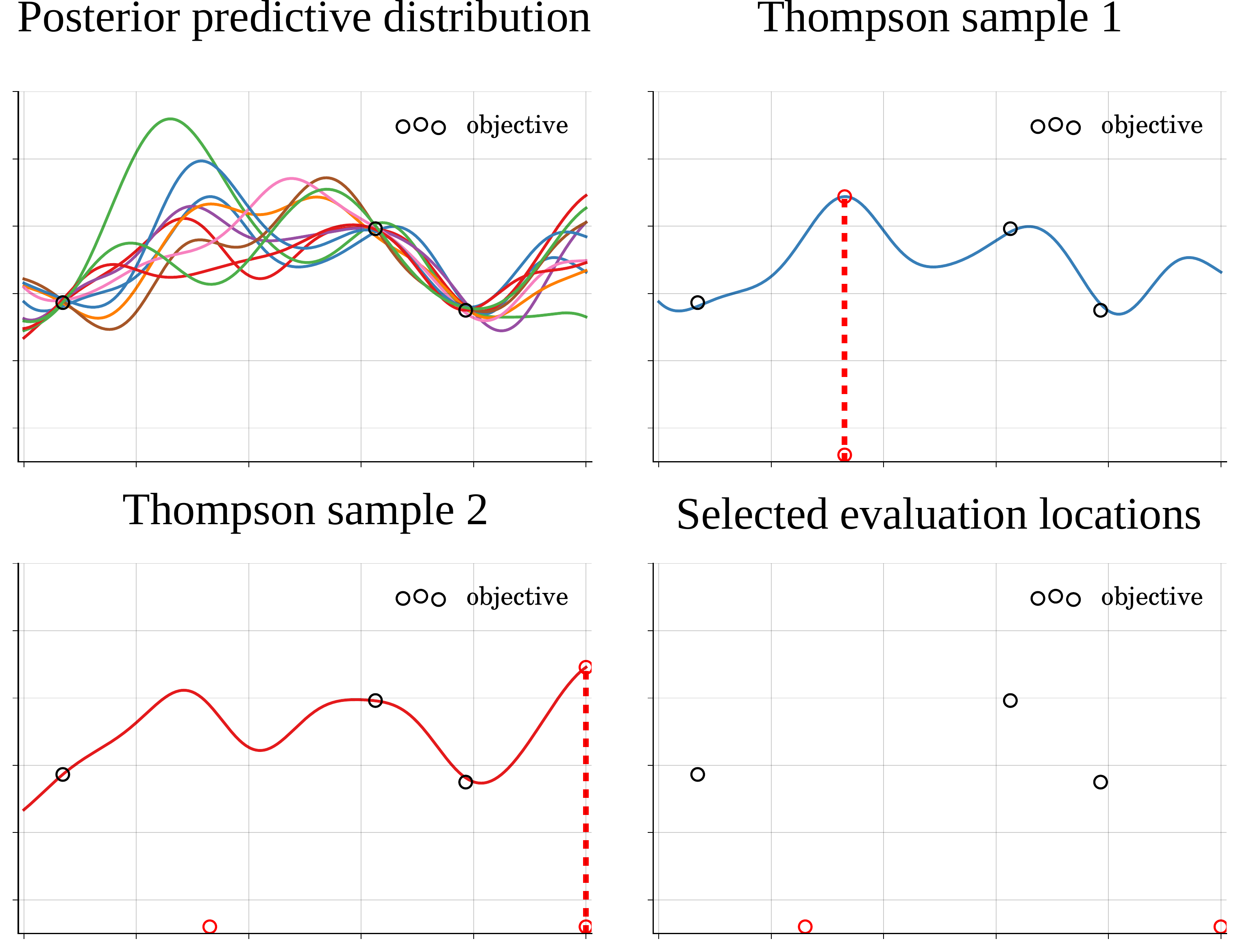}
\caption{Illustration of Thompson sampling and PDTS.}
\vspace{-0.5cm}
\label{fig:illustration_thompson_sampling}
\end{figure}

\vspace{-0.2cm}
\section{BO and Thompson Sampling}
\vspace{-0.1cm}

Let us assume we have a large library of candidate molecules
${\mathcal{M}=\{ m_1,\ldots,m_{|\mathcal{M}|}\}}$. Our goal is to identify a
small subset of elements ${\{m_i\} \subset \mathcal{M}}$ for which the $f(m_i)$ are as high as
possible, with $f$ being an expensive-to-evaluate objective function. The objective $f$ could be, for example, an estimate of the
power-conversion efficiency of organic photovoltaics, as given by expensive
quantum mechanical simulations \cite{ADMA200501717}, and we may want to identify the top 1\% elements in $\mathcal{M}$
according to this score.  

Bayesian optimization methods can be used to identify the inputs that maximize
an expensive objective function $f$ by performing only a reduced number of
function evaluations. For this, BO uses a model to make predictions for the
value of $f$ at new inputs given data from previous evaluations. The next point
to evaluate is then chosen by maximizing an acquisition function that
quantifies the benefit of evaluating the objective at a particular location.

Let $\mathbf{x}_1,\ldots,\mathbf{x}_{|\mathcal{M}|}$ be $D$-dimensional feature
vectors for the molecules in~$\mathcal{M}$ and let~${\mathcal{D}_\mathcal{I}=\{
(\mathbf{x}_i, y_i): i\in I \}}$ be a dataset with information about past
evaluations, where~$I$ is a set with the indices of the molecules already
evaluated,~$\mathbf{x}_i$ is the feature vector for the $i$-th molecule in~$\mathcal{M}$ and~${y_i = f(m_i)}$ is the result of evaluating the objective
function $f$ on that molecule. We assume that the evaluations of $f$ are
noise free, however, the methods described here can be applied to the
case in which the objective evaluations are corrupted with additive Gaussian
noise. BO typically uses a probabilistic model to
describe how the~$y_i$ in~$\mathcal{D}_\mathcal{I}$ are generated as a function of the
corresponding features~$\mathbf{x}_i$ and some model parameters~$\bm \theta$,
that is, the model specifies~$p(y_i|\mathbf{x}_i,\bm\theta)$. 
Given the data~$\mathcal{D}_\mathcal{I}$ and a prior distribution~$p(\bm \theta)$, the
model also specifies a posterior distribution~${p(\bm \theta
|\mathcal{D}_\mathcal{I})\propto p(\bm\theta)\prod_{i \in I}
p(y_i|\mathbf{x}_i,\bm\theta)}$.
The predictive distribution for 
any~${m_j\in\mathcal{M}\setminus \{m_i : i \in I\}}$ is then given by~${p(y_j|\mathbf{x}_j,\mathcal{D}_\mathcal{I})=\int p(y_j|\mathbf{x}_j,\bm\theta)p(\bm \theta
|\mathcal{D}_\mathcal{I})\,d\bm\theta}$.
BO methods use this predictive distribution to
compute an acquisition function (AF) given by

\vspace{-0.4cm}
{\small
\begin{equation}
\alpha(\mathbf{x}_j|\mathcal{D}_\mathcal{I}) =
\mathbf{E}_{p(y_j|\mathbf{x}_j,\mathcal{D}_\mathcal{I})} \left[ U(y_j|\mathbf{x}_j,\mathcal{D}_\mathcal{I}) \right]\,,\label{eq:AF}
\end{equation}}where $U(y_j|\mathbf{x}_j,\mathcal{D}_\mathcal{I})$ is the utility of obtaining value~$y_j$ when evaluating~$f$ at~$m_j$. Eq.~(\ref{eq:AF}) is then maximized with respect to~${j\not \in I}$ to select
the next molecule~$m_j$ on which to evaluate~$f$. The most common choice for the utility
is the improvement:~${U(y_j|\mathbf{x}_j,\mathcal{D}_\mathcal{I})= \max(0, y_j -
y_\star)}$, where~$y_\star$ is equal to the best~$y_i$ in~$\mathcal{D}_\mathcal{I}$.
In this case, Eq.~(\ref{eq:AF}) is called the expected improvement (EI) \cite{jones1998efficient}.
Ideally, the AF should encourage both exploration and exploitation.
For this, the expected utility should increase when $y_j$ takes
high values on average (to exploit), but also when there is high
uncertainty about~$y_j$ (to explore). The EI utility function satisfies these two requirements.

\todo[inline]{This description of Thompson sampling is confusing and doesn't seem correct.  It is needless formalization that does not provide insight.  Thompson sampling is not trying to form a Monte Carlo estimate of the distribution over $y_i$.  It is sampling from the distribution over maxima implied by the posterior.  The exploration in TS does not arise from Monte Carlo variance, but from the true uncertainty associated with the distribution over maxima.  I don't see how this maps into the utility framework.}
Thompson sampling (TS) \cite{Thompson_1933} can be understood as a version of the previous
framework in which the utility function is defined as~${U(y_j|\mathbf{x}_j,\mathcal{D}_\mathcal{I}) = y_j}$ and the expectation in (\ref{eq:AF})
is taken with respect to $p(y_j|\mathbf{x}_j,\bm\theta)$ instead of
$p(y_j|\mathbf{x}_j,\mathcal{D}_\mathcal{I})$, with $\bm \theta$ being a sample
from the posterior $p(\bm \theta|\mathcal{D}_\mathcal{I})$.  That is, when computing the
AF, TS approximates the integral in
$p(y_j|\mathbf{x}_j,\mathcal{D}_\mathcal{I})=\int
p(y_j|\mathbf{x}_j,\bm\theta)p(\bm \theta |\mathcal{D}_\mathcal{I})\,d\bm\theta$ by Monte
Carlo, using a single sample from $p(\bm \theta|\mathcal{D}_\mathcal{I})$ in the
approximation. The TS utility function enforces only exploitation because the
expected utility is insensitive to any variance in $y_j$. Despite this, TS
still enforces exploration because of the variance produced by the Monte Carlo
approximation to $p(y_j|\mathbf{x}_j,\mathcal{D}_\mathcal{I})$. Under TS,
the probability of
evaluating the objective at a particular location matches the probability of
that location being the maximizer of the objective, given the model assumptions
and the data from past evaluations. Algorithm \ref{alg:seq_thompson_sampling}
contains the pseudocode for TS. The plots in the top of Figure
\ref{fig:illustration_thompson_sampling} illustrate how TS works. The top-left
plot shows several samples from a posterior distribution on $f$ induced by
$p(\bm\theta|\mathcal{D}_\mathcal{I})$ since each value of the parameters $\bm \theta$ corresponds to
an associated value of $f$.
Sampling from $p(\bm\theta|\mathcal{D}_\mathcal{I})$ is then equivalent to selecting one of these
samples for $f$. The selected sample 
represents the current AF, which is
optimized in the top-right plot in Figure \ref{fig:illustration_thompson_sampling} 
to select the next evaluation.


\vspace{-0.1cm}
\subsection{Parallel BO}\label{sec:parallelBO}
\vspace{-0.10cm}

So far we have considered the sequential evaluation setting, where BO methods
collect just a single data point in each iteration. However, BO can also be
applied in the parallel setting, which involves choosing a batch of multiple points to
evaluate next in each iteration. For example, when we run
$S$ parallel simulations in a computer cluster and each simulation performs one
evaluation of $f$.
\todo[inline]{One thing that is missing here is clarity on what the nature of the parallelism is.  There are certainly situations where you have a big batch of experiments that are all running in lockstep, in particular in biology with microarrays.  However, a much more common situation is that you're being asked to choose a new place for evaluation, when many other things are currently pending.  Computational chemistry is certainly in this category, since the run times are different for different molecules.  It needs to be clear what kind of parallelism is being discussed.}

\citet{snoek2012practical} describe how to extend sequential BO methods to the
parallel setting. The idea is to select the first evaluation location in the
batch in the same way as in the sequential setting. However, the next evaluation
location  is then selected while the previous one is still pending. In
particular, given a set $K$ with indexes of pending evaluation locations, we
choose a new location in the batch based on the expectation of the AF under all
possible outcomes of the pending evaluations according to the predictions of
the model. Therefore, at any point, the next evaluation location is obtained by
optimizing the AF

\vspace{-0.5cm}
{
\small
\begin{align}
& \alpha_\text{parallel} (\mathbf{x}_j|\mathcal{D}_\mathcal{I},\mathcal{K}) = \nonumber\\
&\hspace{0.9cm} \mathbf{E}_{p(\{y_k\}_{k \in \mathcal{K}}|\{ \mathbf{x}_k\}_{k \in \mathcal{K}},\mathcal{D}_\mathcal{I})}\left[ 
\alpha(\mathbf{x}_j|\mathcal{D}_\mathcal{I} \cup \mathcal{D}_\mathcal{K}) \right]
\,,\label{eq:integratedAF}
\end{align}}where $\mathcal{D}_\mathcal{K}=\{(y_k,\mathbf{x}_k)\}_{k\in \mathcal{K}}$ and
$\alpha(\mathbf{x}_j|\mathcal{D}_\mathcal{I} \cup \mathcal{D}_\mathcal{K})$
is given by (\ref{eq:AF}).
Computing this expression exactly is infeasible in most cases.
\citet{snoek2012practical} propose a Monte Carlo approximation in which the expectation
in the second line is approximated by averaging across a few samples from
the predictive distribution at the pending evaluations,
that is, $p(\{y_k\}_{k \in \mathcal{K}}|\{ \mathbf{x}_k\}_{k \in \mathcal{K}},\mathcal{D}_\mathcal{I})$.
These samples are referred to as \emph{fantasized} data.

\newcommand*\circled[1]{\tikz[baseline=(char.base)]{\node[shape=circle,draw,inner sep=0.4pt] (char) {#1};}} 

This approach for parallel BO has been successfully used to collect small
batches of data (about 10 elements in size), with EI as utility function and
with a Gaussian process as the model for the data \cite{snoek2012practical}.
However, it lacks scalability to large batch sizes, failing when we need to
collect thousands of simultaneous measurements.  The reason for this is the
high computational cost of adding a new evaluation to the current
batch. The corresponding cost includes: \circled{1} sampling the fantasized data, \circled{2} updating
the posterior predictive distribution to
$p(y_j|\mathbf{x}_j,\mathcal{D}_\mathcal{I}\cup \mathcal{D}_\mathcal{K})$,
which is required for evaluating $\alpha(\mathbf{x}_j|\mathcal{D}_\mathcal{I}
\cup \mathcal{D}_\mathcal{K})$, and \circled{3} optimizing the Monte Carlo approximation
to (\ref{eq:integratedAF}). Step \circled{2} can be very expensive when the number of
training points in $\mathcal{D}_\mathcal{I}$ is very large.  This step is
also
considerably challenging when the model does not allow for exact
inference, as it is often the case with Bayesian neural networks.  Step \circled{3} can
also take a very long time when the library of candidate molecules
$\mathcal{M}$ is very large (e.g., when it contains millions of elements)
and among all the remaining molecules we have to find one that maximizes
the AF.

\begin{figure*}
\centering
\includegraphics[width=1.00\textwidth]{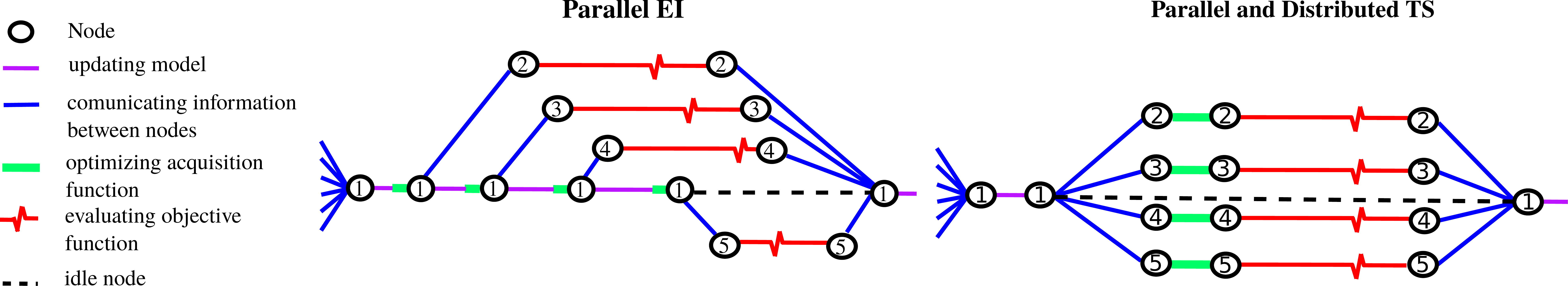}
\caption{A visualization of one iteration of BO using parallel EI as implemented in \cite{snoek2012practical} and PDTS. Note that in PDTS the model is updated once and sample points are acquired independently by the nodes. With parallel EI, the the location of the next sample points is dependent on the location of previous sample points in the batch so these are computed sequentially.
	\todo[inline]{I think this figure could be helpful, but it doesn't really address the situation here in which the experiments take variable amounts of time.  Also, these figures make it clear that it only matters if the red lines are short!}
}
\label{fig:parallel_visualization}
\vspace{-3mm}
\end{figure*}

Despite these difficulties, the biggest disadvantage in this approach for
parallel BO is that it cannot be parallelized since it is a sequential process
in which (\ref{eq:integratedAF}) needs to be iteratively optimized, with each
optimization step having a direct effect on the next one. This prevents this
method from fully exploiting the acceleration provided by multiple processors
in a computer cluster.  \todo{Let's be honest here: this only matters if the ``expensive'' computation is not that expensive relative to the parallelism.  That is, if it takes 1 minute to make a sequential prediction and we have $N$ machines, then once the optimization is going, the bottleneck only arises if the expensive function takes less than $N$ minutes.  In other words, this argument somewhat contradicts the philosophy behind BO.}  The sequential nature of the algorithm is illustrated by
the plot in the left of Figure \ref{fig:parallel_visualization}. In this plot
computer node 1 is controlling the BO process and decides the batch evaluation
locations. Nodes $2,\ldots,5$ then perform the evaluations in parallel. Note
that steps \circled{2} and \circled{3} from the above description have been highlighted in green
and magenta colors. 

In the following section we describe an algorithm \todo[inline]{This algorithm is useful, but no way it's novel.  It's the most obvious thing to do with Thompson sampling in the batch setting.  We can argue that this is a sensible thing to do for chemistry, but I think basically everyone who has thought about TS realizes you can do it trivially in parallel. } for batch BO which can
be implemented in a fully parallel and distributed manner and which,
consequently, can take full advantage of multiple processors in a computer
cluster. This novel method is based on a parallel implementation of the Thompson sampling heuristic.



\begin{algorithm}
   \caption{Parallel and distributed Thompson sampling}\label{alg:thompson_sampling_distributed}
\begin{algorithmic}
   \STATE {\bfseries Input:} initial data $\mathcal{D}_{\mathcal{I}(1)}=\{ \mathbf{x}_i, y_i \}_{i \in \mathcal{I}(1)}$, batch size $S$
   \FOR{$t=1$ {\bfseries to} $T$}
   \STATE  \tikzmark{a} Compute current posterior $p(\bm \theta|\mathcal{D}_{\mathcal{I}(t)})$
   \FOR{$s=1$ {\bfseries to} $S$}
   \STATE Sample $\bm \theta$ from $p(\bm \theta|\mathcal{D}_{\mathcal{I}(t)})$
   \STATE Select $k(s)\leftarrow \text{argmax}_{j \not \in {\mathcal{I}(t)}} \mathbf{E}[y_j|\mathbf{x}_j,\bm \theta]$
   \STATE Collect $y_{k(s)}$ by evaluating $f$ at $\mathbf{x}_{k(s)}$
   \ENDFOR
   \STATE $\mathcal{D}_{\mathcal{I}(t+1)}=\mathcal{D}_{\mathcal{I}(t)}\cup \{\mathbf{x}_{k(s)},y_{k(s)}\}_{s=1}^S$
   \ENDFOR
\begin{tikzpicture}[remember picture, overlay]
   \node
    [draw=blue!80!black, very thick, dotted, rectangle, anchor=north west,
     minimum width=5.9cm,
     minimum height=1.3cm]
    (box) at ($(a) + (0.15, -0.5)$) {};
    \node[text=blue!80!black, rotate=90, anchor=south] at ($(a) + (6.6, -1.16)$) {Executed};
    \node[text=blue!80!black, rotate=90, anchor=south] at ($(a) + (7.0, -1.16)$) {in parallel};
    \node[text=blue!80!black, rotate=90, anchor=south] at ($(a) + (7.3, -1.16)$) {in node $s$};
\end{tikzpicture}
\end{algorithmic}
\end{algorithm}

\vspace{-0.3cm}
\section{Parallel and Distributed Thompson Sampling}
\vspace{-0.1cm}

We present an implementation of the parallel BO method from Section
\ref{sec:parallelBO} based on the Thompson sampling (TS) heuristic. In
particular, we propose to apply to (\ref{eq:integratedAF}) the same 
approximation that TS applied to (\ref{eq:AF}). For this, we choose 
in (\ref{eq:integratedAF}) the same utility function used by TS in the sequential setting, that is,
$U(y_j|\mathbf{x}_j,\mathcal{D}_\mathcal{I} \cup \mathcal{D}_\mathcal{K}) = y_j$.
Then, we approximate the expectation with respect to $\{y_k\}_{k \in \mathcal{K}}$
in (\ref{eq:integratedAF}) by Monte Carlo, averaging across just one sample of
$\{y_k\}_{k \in \mathcal{K}}$
drawn from
$p(\{y_k\}_{k \in \mathcal{K}}|\{ \mathbf{x}_k\}_{k \in \mathcal{K}},\mathcal{D}_\mathcal{I})$.
After that,
$\alpha(\mathbf{x}_j|\mathcal{D}_\mathcal{I} \cup \mathcal{D}_\mathcal{K})$ in
(\ref{eq:integratedAF}) is approximated in the same way as in the sequential
setting by first sampling $\bm \theta$ from $p(\bm
\theta|\mathcal{D}_\mathcal{I}\cup \mathcal{D}_\mathcal{K})$ and then
approximating $p(y_j|\mathbf{x}_j,\mathcal{D}_\mathcal{I}\cup
\mathcal{D}_\mathcal{K})$ with $p(y_j|\mathbf{x}_j,\bm\theta)$.  Importantly,
in this process, sampling first $\{y_k\}_{k \in \mathcal{K}}$ from
$p(\{y_k\}_{k \in \mathcal{K}}|\{ \mathbf{x}_k\}_{k \in
\mathcal{K}},\mathcal{D}_\mathcal{I})$ and then $\bm \theta$ from $p(\bm
\theta|\mathcal{D}_\mathcal{I}\cup \mathcal{D}_\mathcal{K})$ is equivalent to
sampling $\bm \theta$ from just $p(\bm \theta|\mathcal{D}_\mathcal{I})$.  The
reason for this is that updating a posterior distribution with synthetic data
sampled from the model's predictive distribution produces on average the same
initial posterior distribution. The result is that parallel TS with batch size
$S$ is the same as running sequential TS $S$ times without updating the current
posterior $p(\bm \theta|\mathcal{D}_\mathcal{I})$, where each execution of
sequential TS produces one of the evaluation locations in the batch.
Importantly, these executions can be done in distributed manner, with each one running in
parallel in
a different node.

The resulting parallel and distributed TS (PDTS) method is highly scalable and
can be applied to very large batch sizes by running each execution of
sequential TS on the same computer node that will then later evaluate $f$ at
the selected evaluation location.  Algorithm
\ref{alg:thompson_sampling_distributed} contains the pseudocode for PDTS.  The
parallel nature of the algorithm is illustrated by the plot in the right of
Figure \ref{fig:parallel_visualization}. In this plot computer node 1 is
controlling the BO process. To collect four new function evaluations in
parallel, computer node 1 sends the current posterior $p(\bm
\theta|\mathcal{D}_\mathcal{I})$ and $\mathcal{I}$ to nodes $2,\ldots,5$. Each of them
samples then a value for $\bm \theta$ from the posterior and optimizes its own AF given by $\mathbf{E}[y_j|\mathbf{x}_j,\bm \theta]$,
with $j \not \in \mathcal{I}$.
The objective function is evaluated at the selected input and the resulting data is sent back to node 1.
Figure \ref{fig:illustration_thompson_sampling} illustrates how PDTS selects two parallel
evaluation locations. For this, sequential TS is run twice.

The scalability of PDTS makes it a promising method for parallel BO in
high-throughput screening. However, in this type of problem, the optimization
of the AF is done over a discrete set of molecules. Therefore,
whenever we collect a batch of data in parallel with PDTS, several of the
simultaneous executions of sequential TS may choose to evaluate the same
molecule. A central computer node (e.g. the node controlling the BO process)
maintaining a list of molecules currently selected for evaluation can be used
to avoid this problem. In this case, each sequential TS node sends to the central node a ranked
list with the top $S$ (the batch size) molecules according to its AF.  From
this list, the central node then selects the highest ranked molecule that
has not been selected for evaluation before.


\vspace{-0.1cm}
\section{Related Work}
\vspace{-0.1cm}


Ginsbourger et al. \cite{ginsbourger2010kriging} proposed the following framework for parallel BO:
given a set of current observations
$\mathcal{D}_{\mathcal{I}}$ and pending experiments
$\{\mathbf{x}_k\}_{k=1}^\mathcal{K}$, an additional set of fantasies
$\mathcal{D}_\mathcal{K} = \{(\mathbf{x}_k, {y}_k)\}_{k=1}^\mathcal{K}$ can
be assumed to be the result of those pending experiments. A step of Bayesian
optimization can then be performed using the augmented dataset
$\mathcal{D}_\mathcal{I} \cup \mathcal{D}_\mathcal{K}$ and the
acquisition function $\alpha(\mathbf{x}| \mathcal{D}_\mathcal{I} \cup
{\mathcal{D}_\mathcal{K}})$. Two different values are proposed for the fantasies:
the \textit{constant liar}, where ${y}_k = L$ for some constant $L$
and all $k=1\ldots \mathcal{K}$, and the \textit{Kriging believer}, where
${y}_k$ is given by the GP predictive mean at $\mathbf{x}_k$.

\citet{snoek2012practical} compute a Monte Carlo approximation
of the expected acquisition function over potential fantasies
sampled from the model's predictive distribution.
Recent methods have been proposed to
modify the parallel EI procedure to recommend points jointly
\cite{chevalier2013fast,marmin2015differentiating,wang2016parallel}.

\citet{azimi2010batch} describe a procedure called \textit{simulated matching}
whose goal is to propose a batch ${\mathcal{D}_\mathcal{K}}$ of points which is
a good match for the set of samples that a sequential BO policy $\pi$ would
recommend. The authors consider a batch ``good'' if it contains a sample that
yields, with high probability, an objective value close to that of the best
sample produced by a sequential execution of $\pi$.

Several authors have proposed  to extend the \textit{upper confidence bound}
(UCB) heuristic to the parallel setting.  
Since the GP predictive variance depends only on the input location of the observations, \citet{desautels2014parallelizing} propose GP-BUCP acquisition which uses the UCB acquisition with this updated variance.
\citet{contal2013parallel} introduce the Gaussian Process Upper Confidence
Bound with Pure Exploration (GP-UCB-PE). Under this procedure, the first point
is obtained using the standard UCB acquisition function while the remaining
points are sequentially selected to be the ones yielding the highest predictive
variance, while still lying in a region that contains the maximizer with high
probability.

\citet{shah2015parallel} extend the Predictive Entropy Search (PES) heuristic
to the parallel setting (PPES). PPES seeks to recommend a collection of samples
${\mathcal{D}_\mathcal{\mathcal{K}}}$ that yields the greatest reduction in
entropy for the posterior distribution of $\mathbf{x}^\star$, the latent
objective maximizer.  \citet{wu2016parallel} propose the \text{Parallel
Knowledge Gradient Method} which optimizes an acquisition function called the
parallel knowledge gradient (q-KG), a measure of the expected incremental
solution quality after $q$ samples.

An advantage of PDTS over parallel EI and other related methods is that the
approximate marginalization of potential experimental outcomes adds no
extra computational cost to our procedure and so PDTS is highly parallelizable.
Finally, unlike other approaches, PDTS can be applied to a wide
variety of models, such as GPs and Bayesian neural networks, since it only
requires samples from an exact or approximate posterior distribution.



\newcommand{\DistGam}{\text{Gam}}

\vspace{-0.2cm}
\section{Bayesian Neural Networks for High-throughput Screening}
\vspace{-0.1cm}

Neural networks are well-suited for implementing BO on molecules. They produce
state-of-the-art predictions of chemical properties
\cite{Ma_2015,Mayr_2016,ramsundar2015massively} and can be applied to large
data sets by using stochastic optimization 
\cite{bousquet2008tradeoffs}. Typical applications of neural networks focus on
the deterministic prediction scenario. However, 
in large search spaces with multiple local optima (which is
the case when navigating chemical space),
it is desirable to use a probabilistic approach that can produce accurate
estimates of uncertainty for efficient exploration and so, we use
\emph{probabilistic back-propagation} (PBP), a recently-developed technique for
the scalable training of Bayesian neural networks
\cite{hernandez2015probabilistic}. Note that other methods for approximate
inference in Bayesian neural networks could have been chosen as well
\cite{BlundellCKW15,SnoekRSKSSPPA15,GalG16}. We prefer PBP because it is fast
and it does not require the tuning of hyper-parameters such as learning rates
or regularization constants \cite{hernandez2015probabilistic}.

Given a dataset $\mathcal{D}_\mathcal{I} = \{ (\mathbf{x}_i, y_i) \}_{i \in\mathcal{I}}$, we
assume that ${y_i = f(\mathbf{x}_i;\mathcal{W}) + \epsilon_i}$, where $f(\cdot
;\mathcal{W})$ is the output of a neural network with weights $\mathcal{W}$.
The network output is corrupted with additive noise variables ${\epsilon_i \sim
\mathcal{N}(0,\gamma^{-1})}$. The network has~$L$ layers, with $V_l$ hidden
units in layer $l$, and ${\mathcal{W} = \{ \mathbf{W}_l \}_{l=1}^L}$ is the
collection of $V_l \times (V_{l-1}+1)$ synaptic weight matrices. The $+1$ is
introduced here to account for the additional per-layer biases. The activation
functions for the hidden layers are rectifiers: ${\varphi(x) = \max(x,0)}$. 

The likelihood for the network weights~$\mathcal{W}$ and the noise
precision~$\gamma$ is 

\vspace{-0.5cm}
{\small
\begin{align}
p(\{y_i\}_{i\in|\mathcal{I}}|\mathcal{W},\{\mathbf{x}_i\}_{i\in \mathcal{I}},\gamma) 
&= \prod_{i\in \mathcal{I}}\mathcal{N}(y_i | f(\mathbf{x}_i;\mathcal{W}),\gamma^{-1})\,.\nonumber
\end{align}
}We specify a Gaussian prior distribution
for each entry in each of the weight matrices in $\mathcal{W}$:
\begin{align}
p(\mathcal{W}|\lambda) &= \prod_{l=1}^L \prod_{k=1}^{V_l} \prod_{j=1}^{V_{l-1}+1} \mathcal{N}(w_{kj,l}|0,\lambda^{-1})\,,\label{eq:prior_weights}
\end{align}
where $w_{kj,l}$ is the entry in the $k$-th row and $j$-th column
of $\mathbf{W}_l$ and $\lambda$ is a precision parameter. The hyper-prior
for~$\lambda$ is gamma: $p(\lambda) = \DistGam(\lambda|\alpha_0^\lambda,\beta_0^\lambda)$ 
with shape $\alpha^\lambda_0 = 6$ and inverse scale $\beta^\lambda_0 = 6$.
This relatively low value for the shape and inverse scale parameters makes this prior weakly-informative. 
The prior for the noise precision $\gamma$ is also gamma: $p(\gamma) =
\DistGam(\gamma|\alpha_0^{\gamma},\beta_0^{\gamma})$. We assume that the
$y_i$ have been normalized to have unit variance and, as above, we
fix ${\alpha^{\gamma}_0 = 6}$ and~${\beta^{\gamma}_0 = 6}$.

The exact
computation of the posterior distribution for the model parameters $p(\mathcal{W},\gamma, \lambda|\mathcal{D}_\mathcal{I})$ is not tractable in most cases. 
PBP approximates the intractable posterior on $\mathcal{W}$, $\gamma$ and $\lambda$ with the tractable approximation

\vspace{-0.5cm}
{
\small
\begin{align}
q(\mathcal{W},\gamma, \lambda) = & \left[ \prod_{l=1}^L\! \prod_{k=1}^{V_l}\! 
\prod_{j=1}^{V_{l\!-\!1}\!+\!1} \mathcal{N}(w_{kj,l}| m_{kj,l},v_{kj,l})\right ]\nonumber\\
& \DistGam(\gamma \,|\, \alpha^\gamma, \beta^\gamma)
\DistGam(\lambda \,|\, \alpha^\lambda, \beta^\lambda)\,,\label{eq:posterior_approximation}
\end{align}}whose parameters are tuned by iteratively running an assumed density filtering
(ADF) algorithm over the training data \cite{Opper1998}. The main operation in
PBP is the update of the mean and variance parameters of $q$, that is, the
$m_{kj,l}$ and $v_{kj,l}$ in (\ref{eq:posterior_approximation}), after
processing each data point $\{(\mathbf{x}_i,y_i)\}$.  For this, PBP matches
moments between the new $q$ and the product of the old $q$ with the
corresponding likelihood factor $\mathcal{N}(y_i \,|\,
f(\mathbf{x}_i;\mathcal{W}),\gamma^{-1})$. The matching of moments for the
distributions on the weights is achieved by using well-known Gaussian ADF
updates, see equations 5.12 and 5.1 in \cite{minka2001family}.

To compute the ADF updates, PBP finds a Gaussian approximation to the distribution
of the network output $f(\mathbf{x}_i;\mathcal{W})$ when $\mathcal{W} \sim q$.
This is achieved by doing a forward pass of 
 $\mathbf{x}_i$ through the network, with the weights $\mathcal{W}$ being
randomly sampled from $q$. In this forward pass the non-Gaussian distributions
followed by the output of the neurons are approximated with Gaussians that have the same
means and variances as the original distributions. This is a Gaussian approximation by moment matching. 
We refer the reader to
\citet{hernandez2015probabilistic} for full details on PBP. 

After several ADF iterations over the data by PBP, we can then make predictions
for the unknown target variable $y_\star$ associated with a new feature vector
$\mathbf{x}_\star$. For this, we obtain a Gaussian approximation to
$f(\mathbf{x}_\star;\mathcal{W})$ when $\mathcal{W}\sim q$ by applying the
forward pass process described above.

To implement TS, as described in Algorithm \ref{alg:seq_thompson_sampling}, we
first sample the model parameters $\bm \theta$ from the posterior
$p(\bm \theta|\mathcal{D}_{\mathcal{I}})$ and then optimize the AF
given by $\mathbf{E}[y_j|\mathbf{x}_j,\bm \theta]$, with $j \not \in {\mathcal{I}}$.
When the model is a Bayesian neural network trained with PBP, the corresponding operations
are sampling $\mathcal{W}$ from $q$ and then optimizing the AF
given by
$f(\mathbf{x}_j;\mathcal{W})$, with $j \not \in {\mathcal{I}}$. This last step
requires the use of a deterministic neural network, with weight values given by the
posterior sample from $q$, to make predictions on all the molecules that have
not been evaluated yet. Then, the molecule with highest predictive value is 
selected for the next evaluation.


\section{Experiments with GPs and Parallel EI}

\begin{figure*}[h!]


	 	\includegraphics[width=1.00\textwidth]{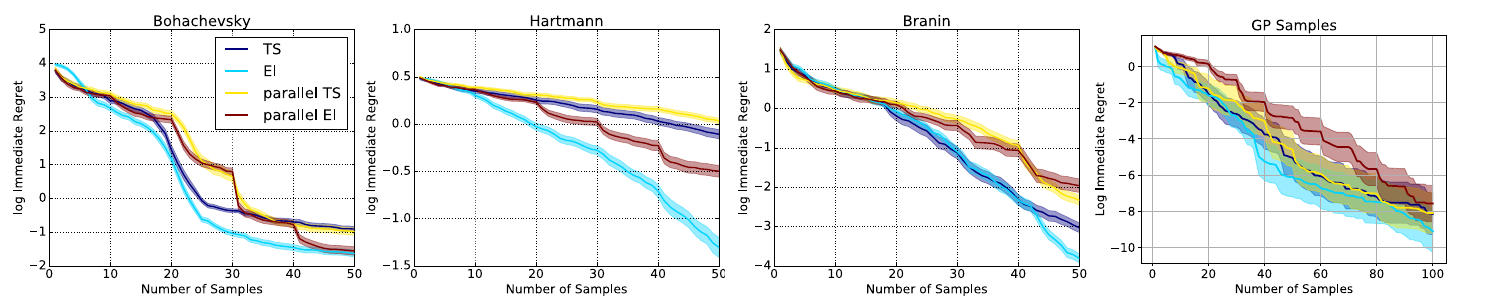}



\caption{Immediate regret in experiments with GPs, using TS,
EI, PDTS and parallel EI for optimizing synthetic functions (first 3 plots) and 
functions sampled from a GP prior (fourth plot).}\label{fig:gp_experiments}
\vspace{-4mm}
\end{figure*}

We first compare the performance of our parallel and distributed Thompson
sampling (PDTS) algorithm with the most popular approach for parallel BO: the
parallel EI method from Section \ref{sec:parallelBO}. Existing implementations
of parallel EI such as
spearmint\footnote{\url{https://github.com/HIPS/Spearmint}} use a Gaussian
process (GP) model for the objective function. To compare with these methods, we
also adopt a GP as the model in PDTS. Note that parallel EI cannot
scale to the large batch sizes used in high-throughput screening. Therefore,
we consider here only parallel optimization problems with small batch sizes and
synthetic objective functions.
Besides PDTS and parallel EI, we also analyze the performance of the sequential
versions of these algorithms: TS and EI.

To implement Thompson sampling (TS) with a GP model, we approximate the
non-parametric GP with a parametric approximation based on random features, as
described in the supplementary material of \cite{hernandez2014predictive}. For
the experiments, we consider a cluster with 11 nodes: one central node for
controlling the BO process and 10 additional nodes for parallel evaluations. We
assume that all objective evaluations take a very large amount of time and
that the cost of training the GPs and recomputing and optimizing the AF is
negligible in comparison. Thus, in practice, we perform these experiments
in a sequential (non-parallel) fashion with the GP model being updated only in
blocks of 10 consecutive data points at a time.

As objective functions we consider the two dimensional Bohachevsky and
Branin-Hoo functions and the six dimensional Hartmann function, all available
in Benchfunk\footnote{\url{https://github.com/mwhoffman/benchfunk}}. We also
consider the optimization of functions sampled from the GP prior over the 2D unit square using a
squared exponential covariance function with fixed 0.1 length scale. After each objective evaluation, we compute the immediate
regret (IR), which we define as the difference between the best objective value
obtained so far and the minimum value of the objective function. The measurement
noise is zero in these experiments.

Figure \ref{fig:gp_experiments} reports mean and standard errors for the
logarithm of the best IR seen so far, averaged across 50 repetitions of the experiments. 
In the plots, the horizontal axis
shows the number of function evaluations performed so far.  Note that in these
experiments TS and EI update their GPs once per sample, while PDTS and parallel
EI update only every 10 samples. 
Figure \ref{fig:gp_experiments} shows that EI is
better than TS in most cases, although the differences between these two
methods are small in the Branin-Hoo function.  However, EI is considerably much
better than TS in Hartmann. The reason for this is that in Hartmann there are
multiple equivalent global minima and TS tends to explore all of them. EI is by contrast
more exploitative and focuses on evaluating the objective around only one of
the minima. 
The differences between parallel EI and PDTS are much
smaller, with both obtaining very similar results.
The 
exception is again Hartmann, where parallel EI is much better than PDTS,
probably because PDTS is more explorative than
parallel EI. Interestingly, PDTS performs better than parallel EI
on the random samples from the GP prior, although parallel EI eventually catches up.

These results indicate that PDTS performs in practice very similarly to parallel EI,
one of the most popular methods for parallel BO.




\section{Experiments with Molecule Data Sets}\label{sec:data_sets}

We describe the molecule data sets used in our experiments. The input features for
all molecules are 512-bit Morgan circular fingerprints \cite{Rogers_2010},
calculated with a bond radius of 2, and derived from the canonical SMILES as
implemented in the RDkit package \cite{rdkit}.

\textbf{Harvard Clean Energy Project}: The Clean Energy Project is the world's
largest materials high-throughput virtual screening effort
\cite{Hachmann_2014,Hachmann_2011}, and has scanned more than 3.5 million
molecules to find those with high power conversion efficiency (PCE) using
quantum-chemical techniques, taking over 30,000 years of CPU time. The target
value within this data set is the power conversion efficiency (PCE), which is
calculated for the 2.3 million publicly released molecules, using the Scharber
model \cite{Dennler_2008} and frontier orbitals calculated at the BP86
\cite{Perdew_1986,Becke_1993} \/ def2-SVP \cite{Weigend_2005} level of theory.

\textbf{Dose-Response Data Set}: These data sets were obtained from the
NCI-cancer database \cite{_nci_}.  The dose-response target value has a
potential range of -100 to 100, and reports a percentage cell growth relative
to a no-drug control.  Thus, a value of +40 would correspond to a 60\% growth
inhibition and a value of -40 would correspond to 40\% lethality.  Molecules
with a positive value for the dose-response are known as inhibitors, molecules
with a score less than 0 have a cytotoxic effect. Results against the NCI-H23
cell line were taken against a constant log-concentration of -8.00M and where
multiple identical conditions were present in the data an average was used for
the target variables. In this data set we are interested in finding molecules
with  smaller values of the target variable.

\textbf{Malaria Data Set}: The Malaria data set was taken from the \textit{P.
falciparum} whole cell screening derived by combining the GSK TCAMS data set,
the Novatis-GNF Malaria Box data set and the St Jude's Research Hospital data
set, as released through the Medicines for Malaria Venture website
\cite{Spangenberg_2013}. The target variable is the EC50 value, which is
defined as the concentration of the drug which gives half maximal response.
Much like the Dose response data set, the focus here is on minimization: the
lower the concentration, the stronger the drug.

\begin{figure*}[h!]
\begin{center}
\includegraphics[width=1\textwidth]{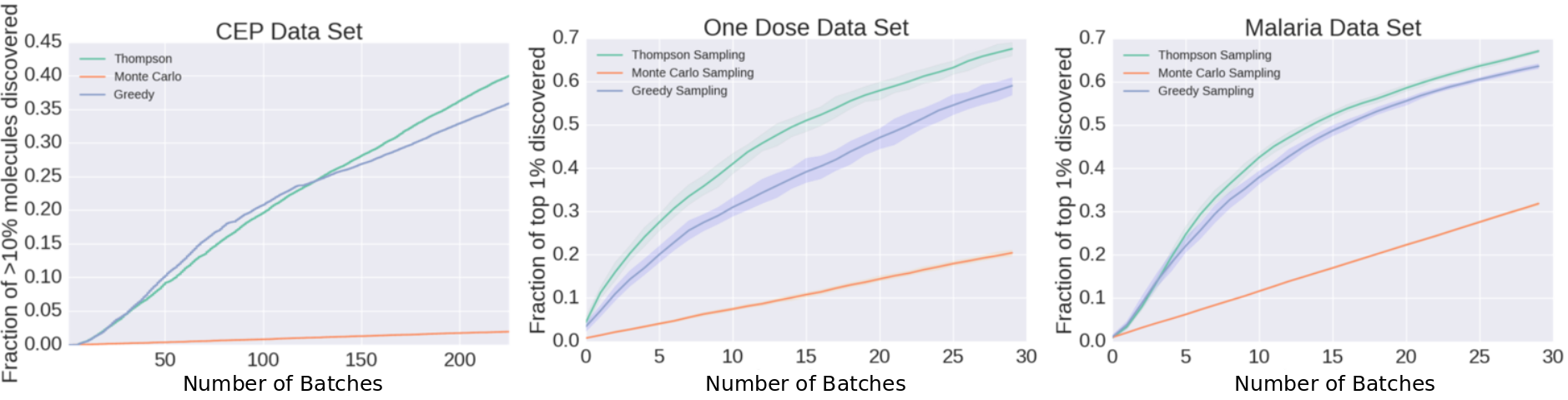}

\vspace{-3mm}
\caption{{Recall obtained by PDTS on each data set. 
For the CEP
data, the recall for molecules with a PCE $>10\%$ is reported, whilst for
One-dose and Malaria we report the recall for the molecules in the top 1\%. In
addition to the Monte Carlo sampling baseline, we also include results for a
greedy sampling approach, in which there is no exploration, and the molecules
are chosen according to the mean of the predictive distribution given by PBP.
The overall lower performance of this greedy strategy illustrates the
importance of exploration in this type of problems. 
\label{fig:thompson_1pc}
}}
\end{center}
\vspace{-4mm}
\end{figure*}

\subsection{Results}\label{sec:thompson_sampling}

We evaluate the gains produced by PDTS in experiments
simulating a high throughput virtual screening setting. In these experiments,
we sequentially sample molecules from libraries of candidate molecules given by
the data sets from Section \ref{sec:data_sets}. After each sampling step, we
calculate the 1\% recall, that is, the fraction of the top 1\% of molecules
from the original library that are found among the sampled ones. For the CEP
data, we compute recall by focusing on molecules with 
PCE larger than 10\%. In all data sets, each sampling step
involves selecting a batch of molecules among those that have not been sampled
so far. In the Malaria and One-dose data sets we use batches of size 200. These
data sets each contain about 20,000 molecules. By contrast, the CEP data set
contains 2 million molecules. In this latter case, we use batches of size 500.
We use Bayesian neural networks with one hidden layer
and 100 hidden units.

We compare the performance of PDTS with two baselines. The first one,
\emph{greedy}, is a sampling strategy that only considers exploitation and does
not perform any exploration. We implement this approach by selecting molecules
according to the average of the probabilistic predictions generated by PBP.
That is, the greedy approach ignores any variance in the predictions of the
Bayesian neural network and generates batches by just ranking molecules
according to the mean of the predictive distribution given by PBP. The second
baseline is a Monte Carlo approach in which the batches of molecules are
selected uniformly at random. These two baselines are comparable to PDTS in that
they can be easily implemented in a large scale setting in which the library of
candidate molecules contains millions of elements and data is sampled using
large batch sizes.

In the Malaria and One-dose data sets, we average across 50 different
realizations of the experiments. This is not possible in the CEP data set,
which is 100 times larger than the two other data sets. In the CEP case, we
report results for a single realization of the experiment (in a second
realization we obtained similar results).  Figure \ref{fig:thompson_1pc} shows
the recall obtained by each method in the molecule data sets.
PDTS significantly outperforms the Monte Carlo
approach, and also offers better performance than greedy sampling. This shows
the importance of building in exploration into the sampling strategy, rather
than relying on purely exploitative methods. The greedy approach performs best
in the CEP data set. In this case, the greedy strategy initially finds better
molecules than PDTS, but after a while PDTS overtakes, probably because a promising
area of chemical space initially discovered by the greedy approach starts to
become exhausted. 

The previous results allow us to consider the savings produced by BO.
In the CEP data set, PDTS achieves about 20 times higher
recall values than the Monte Carlo approach, which is comparable to the
exhaustive enumeration that was used to collect the CEP data. We estimate that,
with BO, the CEP virtual screening process would have
taken 1,500 CPU years instead of the 30,000 that were actually used. Regarding
the One-dose and Malaria data sets, PDTS can locate in both sets about 70\% of
the top 1\% molecules by sampling approximately 6,000 molecules. By contrast,
the Monte Carlo approach would require sampling 14,000 molecules. This
represents a significant reduction in the discovery time for new therapeutic
molecules and savings in the economic costs associated with molecule synthesis
and testing.


\vspace{-0.1cm}
\subsection{Comparison with $\epsilon$-greedy Approaches}
\vspace{-0.1cm}

We can easily modify the greedy baseline from the previous section to include
some amount of exploration by replacing a small fraction of the
molecules in each batch with molecules chosen uniformly at random. This
approach is often called $\epsilon$-greedy \cite{watkins1989learning}, where
the variable $\epsilon$ indicates the fraction of molecules that are sampled
uniformly at random. The disadvantage of the $\epsilon$-greedy approach is that
it requires the tuning of $\epsilon$ to the problem of interest
whereas the amount of exploration is automatically set by PDTS.

\begin{table}
\centering
\caption{Average rank and standard errors by each method.}\label{table1}
\begin{tabular}{lr@{$\pm$}l}
\hline
\bf{Method}& \multicolumn{2}{c}{\bf{Rank}}\\
\hline
$\epsilon = 0.01$ & 3.42 & 0.28 \\
$\epsilon = 0.025$ & 3.02 & 0.25 \\
$\epsilon = 0.05$ & 2.86 & 0.23 \\
$\epsilon = 0.075$ & 3.20 & 0.26 \\
PDTS & \bf{2.51} & \bf{0.20} \\
\hline
\vspace{-8mm}
\end{tabular}
\end{table}


We compared PDTS with different versions of $\epsilon$-greedy in the same way as
above, using $\epsilon = 0.01, 0.025, 0.05$ and $0.075$. The experiments with
the One-dose and the Malaria data sets are similar to the ones done before.
However, we now sub-sample the CEP data set to be able to average across 50
different realizations of the experiment: we choose 4,000 molecules uniformly
at random and then collect data in batches of size 50 across 50 different
repetitions of the screening process. We compute the average rank obtained by
each method across the $3\times 50 = 150$ simulated screening experiments. A
ranking equal to 1 indicates that the method always obtains the highest recall
at the end of the experiment, while a ranking equal to 5 indicates that the
method always obtains the worst recall value. Table \ref{table1} shows that the
lowest average rank is obtained by PDTS, which achieves better
exploration-exploitation trade-offs than the $\epsilon$-greedy approaches.


\vspace{-0.1cm}
\section{Conclusions}
\vspace{-0.1cm}

We have presented a Parallel and Distributed implementation of Thompson
Sampling (PDTS), a highly scalable method for parallel Bayesian
optimization. PDTS can be applied when scalability limits the
applicability of competing approaches. We have evaluated the performance of
PDTS in experiments with both Gaussian process and probabilistic neural
networks. We show that PDTS compares favorably with parallel EI in problems with
small batch sizes. We also demonstrate the effectiveness of PDTS 
on large scale real world applications that involve searching chemical space
for new molecules wit improved properties. We show that PDTS outperforms other scalable
approaches on these applications, in particular, a greedy search strategy,
$\epsilon$-greedy approaches and a random search method.

\subsection*{Acknowledgements}

J.M.H.L. acknowledges support from the Rafael del Pino Foundation.
The authors thank Ryan P. Adams for useful discussions.
A.A.-G. and E.O.P.-K. acknowledge the Department of Energy Program on Theory and modeling through grant {DE-SC0008733}.

\end{document}